\documentclass[runningheads]{llncs}
\usepackage[T1]{fontenc}
\usepackage{esvect}
\usepackage{bbm}
\usepackage{amsmath}
\usepackage{amssymb}
\usepackage{graphicx}
\usepackage{booktabs}
\begin{document}
\title{BRIDGE the Gap: Mitigating Bias Amplification in Automated Scoring of English Language Learners via Inter-group Data Augmentation}

\titlerunning{Mitigating Bias Amplification via Inter-group Data Augmentation}
%
%

\author{Yun Wang\inst{1}\textsuperscript{*}\orcidID{0009-0004-6611-0752}\and
Xuansheng Wu\inst{1}\textsuperscript{*}\orcidID{0000-0002-7816-7658} \and
Jingyuan Huang\inst{1}\orcidID{0009-0002-9935-7685} \and
Lei Liu\inst{2}\orcidID{0000-0002-8327-2700} \and
Xiaoming Zhai\inst{3}\orcidID{0000-0003-4519-1931}\and
Ninghao Liu\inst{4}\orcidID{0000-0002-9170-2424}
}
\authorrunning{Y. Wang et al.}
%
\institute{School of Computing, University of Georgia, Athens, GA, USA \and
Educational Testing Service, Princeton, NJ, USA \and
AI4STEM Education Center, University of Georgia, Athens, GA, USA \and
The Hong Kong Polytechnic University, Hong Kong, China\\
\email{\{Yun.Wang1,Xuansheng.Wu,Jingyuan.Huang,Xiaoming.Zhai\}@uga.edu, lliu001@ets.org, ninghliu@polyu.edu.hk}}
\maketitle              
\begingroup
\renewcommand{\thefootnote}{\fnsymbol{footnote}}
\setcounter{footnote}{0}
\footnotetext{* These authors contributed equally.}
\endgroup
\begin{abstract}
In educational assessment, automated scoring systems increasingly rely on deep learning and large language models (LLMs). However, these systems face significant risks of bias amplification, where model prediction gaps between student groups become larger than those observed in training data. This issue is especially severe for underrepresented groups such as English Language Learners (ELLs), as models may inherit and further magnify existing disparities in the data. We identify that this issue is closely tied to representation bias: the scarcity of minority (high-scoring ELL) samples makes models trained with empirical risk minimization favor majority (non-ELL) linguistic patterns. Consequently, models tend to under-predict ELL students who even demonstrate comparable domain knowledge but use different linguistic patterns, thereby undermining fairness in automated scoring.
To mitigate this, we propose BRIDGE, a \textbf{B}ias-\textbf{R}educing \textbf{I}nter-group \textbf{D}ata \textbf{GE}neration framework designed for low-resource assessment settings. Instead of relying on the limited minority samples, BRIDGE synthesizes high-scoring ELL samples by ``pasting'' construct-relevant (i.e., rubric-aligned knowledge and evidence) content from abundant high-scoring non-ELL samples into authentic ELL linguistic patterns. We further introduce a discriminator model to ensure the quality of synthetic samples. Experiments on California Science Test datasets demonstrate that BRIDGE effectively reduces prediction bias for high-scoring ELL students while maintaining overall scoring performance. Notably, our method achieves fairness gains comparable to using additional real human data, offering a cost-effective solution for ensuring equitable scoring in large-scale assessments.

\end{abstract}
\section{Introduction}
Automated scoring systems powered by deep learning and large language models (LLMs) have been increasingly deployed in educational assessment~\cite{emirtekin2025large,wang2026autoscore,xu2024large}, demonstrating strong performance and promising scalability for large-scale evaluations~\cite{dikli2014automated,zhai2023large}. These systems are inherently data-driven, with predictive performance heavily dependent on the quality and representativeness of the training data~\cite{chaudhari2024deep,9779091}. However, such reliance creates a critical vulnerability in which imbalances in the training data are not just inherited but potentially amplified in model predictions~\cite{mehrabi2021survey,schwartz2022towards}. This phenomenon, known as \emph{bias amplification}, can widen group-level performance disparities in predicted scores beyond those warranted by actual proficiency differences~\cite{andersen2025algorithmic,palermo2022rater,plecko2024mind}. It thereby undermines the fairness of automated scores in educational settings, particularly for underrepresented subgroups such as English Language Learners (ELLs)~\cite{wilson2024validity}.

Such bias amplification can arise from the interaction between data-level \emph{representation bias} and the empirical risk minimization (ERM) objective used to train deep models~\cite{hashimoto2018fairness,mehrabi2021survey}. Representation bias occurs when certain parts of the target population are severely underrepresented in the training data~\cite{shahbazi2023representation}. In educational assessment datasets, ELLs constitute only a small proportion of the student population, with high-scoring ELL responses being even rarer~\cite{welch2025response,wilson2024validity}. This places high-performing ELL students in an extremely sparsely supervised region of the data distribution. Under the ERM paradigm, models prioritize majority patterns to minimize average loss, effectively down-weighting learning signals from underrepresented subgroups~\cite{hashimoto2018fairness,sagawa2019distributionally}. As a result, models may primarily learn scoring criteria from non-ELL exemplars, potentially assigning lower scores to ELL responses that employ different but valid linguistic strategies, even when their responses demonstrate comparable domain knowledge. This systematic underprediction can distort predictions for high-performing ELL students and widen group-level disparities beyond those present in the training data.

To mitigate bias amplification and its downstream fairness impacts, a natural strategy is to enlarge the training data for underrepresented subgroups~\cite{shahbazi2023representation}. Given that manually collecting new high-scoring ELL responses is difficult in practice, targeted synthetic data generation becomes a practical alternative. However, existing augmentation methods offer limited leverage when the target subgroup is extremely underrepresented, precisely the regime where augmentation is most needed. Most existing methods implicitly assume access to a non-trivial pool of subgroup examples to enable resampling or rewriting~\cite{alkhawaldeh2023challenges,chai2025text,chen2022dataaugmentationintentclassification}. When such samples are scarce, classical resampling approaches can lead models to overfit the few minority samples rather than learning generalizable patterns~\cite{alkhawaldeh2023challenges,park2022majority}. Similarly, simple paraphrasing methods, including those based on LLMs, can increase apparent sample diversity, but they face a trade-off between diversity and fidelity~\cite{chai2025text,chen2022dataaugmentationintentclassification,li2026less}. Mild paraphrases often yield near-duplicate responses with limited additional learning signal, whereas aggressive rewrites can drift in meaning and inadvertently alter rubric-relevant content or normalize away subgroup-specific language patterns~\cite{sourati2025homogenizing}. Moreover, such paraphrastic transformations are typically weakly constrained, making it hard to audit which aspects of the response are preserved or changed. This lack of transparency raises concerns about controllability and interpretability in assessment settings.

\begin{figure}[t]
\includegraphics[width=\textwidth]{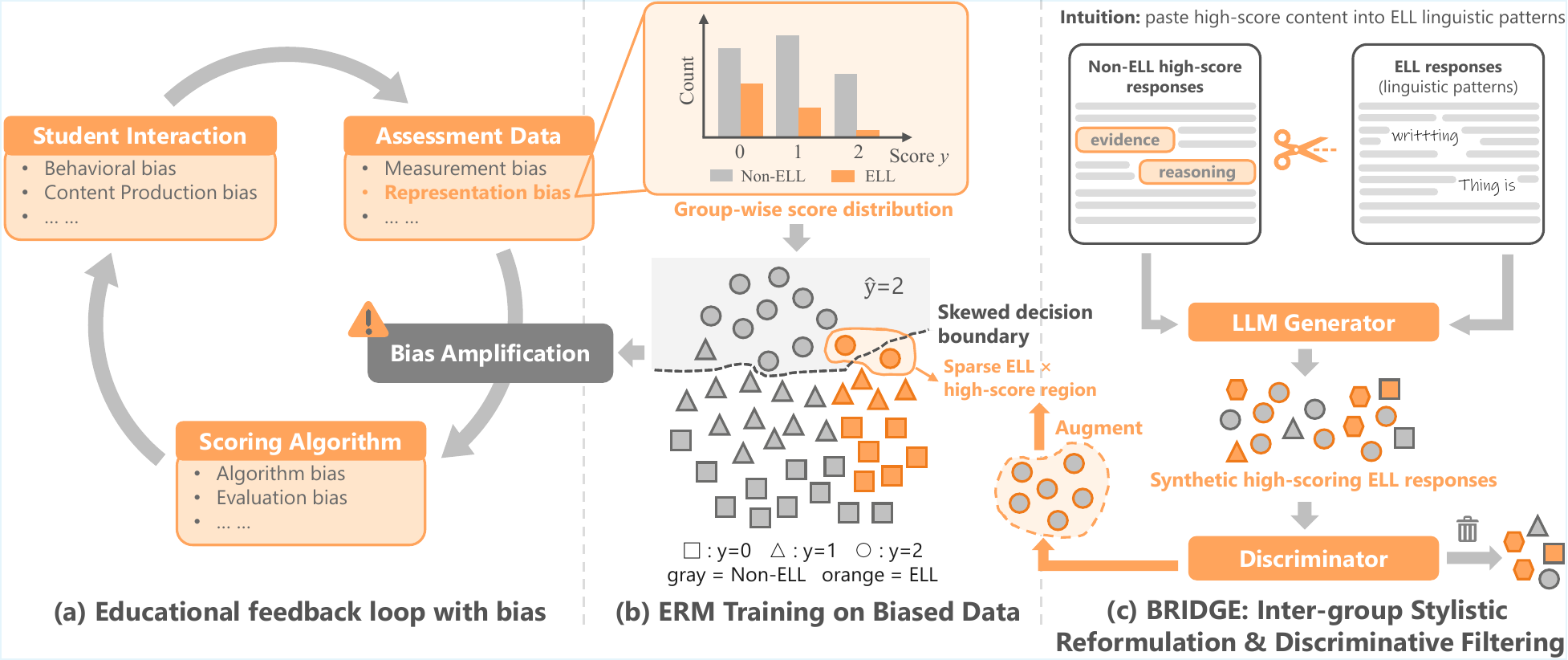}
\caption{
Overview of the bias amplification loop and the BRIDGE framework.
\textbf{(a) Bias Propagation:} Representation bias in training data creates a feedback loop that reinforces educational disparities.
\textbf{(b) ERM Vulnerability:} Under ERM, the decision boundary for high scores is skewed by the majority non-ELL group, causing systematic underprediction of the sparse ELL high-score subgroup.
\textbf{(c) BRIDGE Framework:} Our approach mitigates this by performing inter-group stylistic reformulation, extracting construct-relevant content from high-scoring non-ELL samples and injecting ELL-specific linguistic patterns, followed by a discriminator for quality control.
}
\label{architecture}
\end{figure}

To address these challenges, we propose BRIDGE, an inter-group data augmentation framework designed for extreme underrepresentation scenarios, as illustrated in Fig.~\ref{architecture}. Unlike resampling or paraphrasing methods whose effectiveness depends on having sufficient examples from the target subgroup, BRIDGE can generate high-quality training data even when the subgroup contains only a handful of samples. Our key insight, grounded in a modular view of student responses~\cite{abedi2006psychometric,messick1987validity}, is that semantic content indicates proficiency (i.e., high scores) while linguistic patterns reflect group status (e.g., ELL vs. non-ELL). Thus, although high-scoring ELL responses may be nearly absent, we have access to two abundant and complementary resources: high-scoring non-ELL responses that demonstrate group-invariant construct-relevant content, and other ELL responses that exhibit group-specific linguistic patterns. BRIDGE leverages construct-relevant content from high-scoring non-ELL responses and reformulates it within linguistic patterns characteristic of ELL students. This process generates synthetic responses that adhere to the scoring rubric while reflecting the linguistic characteristics of ELL, offering a more controlled and interpretable augmentation procedure than generic paraphrasing. To ensure the quality of the synthetic data, we further train a discriminator model to automatically filter candidate responses. The main contributions of this work are as follows:

\begin{enumerate}
    \item Revealing and quantifying bias amplification in automated scoring models on high-scoring ELLs, with theoretical analysis from an ERM perspective.
    \item Proposing BRIDGE, an inter-group data augmentation framework that mitigates extreme underrepresentation by blending high-score content with subgroup specific linguistic patterns without requiring additional real data.
    \item Demonstrating on real educational datasets that BRIDGE reduces prediction bias while maintaining scoring performance, achieving fairness gains comparable to real-data augmentation.
\end{enumerate}

\section{Bias Amplification in Automated Scoring}
\label{sec:biasamp}
This section formally defines the automated scoring task and the associated phenomenon of bias amplification. Section~\ref{sec:erm-problem} formalizes the scoring task and introduces the Empirical Risk Minimization (ERM) paradigm. Section~\ref{sec:erm-imbalance} introduces the concept of representation bias and analyzes how it interacts with ERM to impede learning from underrepresented groups. Finally, Section~\ref{sec:rep-to-ba} characterizes the mechanism of bias amplification and defines a metric to quantify it.

\subsection{Problem Formulation}
\label{sec:erm-problem}

We consider an automated scoring task where each example consists of
a student response $x \in \mathcal{X}$,
a holistic score $y \in \mathcal{Y}=\{0,1,\ldots,K\}$ assigned by human raters,
and a binary group label $g \in \mathcal{G}=\{\text{EL}, \text{NE}\}$ indicating the student's English learner status,
where $K$ is the maximum score level, EL refers to ELLs, and NE to non-ELLs.
Let $P$ denote the data distribution over
$\mathcal{X} \times \mathcal{Y} \times \mathcal{G}$.
The training set consists of $n$ i.i.d.\ examples
$D = \{(x_i,y_i,g_i)\}_{i=1}^n$ drawn from $P$.
We write $n_{\text{EL}}$ and $n_{\text{NE}}$ for the number of ELL and non-ELL responses in the training data.

A scoring model $f_\theta : \mathcal{X} \rightarrow \mathbb{R}$ maps an arbitrary response $x$ to a predicted score $\hat{y} = f_\theta(x)$ that approximates the human annotation $y$, where $\theta$ denotes the model parameters.
The parameters are learned by minimizing the empirical risk over the training set:
\begin{equation}
  \label{eq:erm}
  \hat{R}(f_\theta)
    = \frac{1}{n} \sum_{i=1}^{n}
      \ell\!\bigl(f_\theta(x_i), y_i\bigr),
\end{equation}
where $\ell$ denotes a standard loss function. This objective corresponds to the usual ERM paradigm~\cite{vapnik1991principles}.

From a fairness perspective, we are particularly interested in performance for high-scoring ELL students. Let $\tau$ denote a high-score threshold (for example, the top score level). We define the target subgroup as $\mathcal{S}_{\text{EL,high}} = \{ i \mid g_i = \text{EL},\, y_i \ge \tau \}$. Under the ERM objective in Equation~\eqref{eq:erm}, all training examples are weighted equally in the loss. When $\lvert \mathcal{S}_{\text{EL,high}} \rvert$ is very small, the contribution of this subgroup to the overall objective becomes negligible.

\subsection{Representation Bias under ERM}
\label{sec:erm-imbalance}

Building on the ERM formulation in Section~\ref{sec:erm-problem}, we now analyze how representation bias in the training distribution interacts with ERM and impedes the model's learning from underrepresented groups. Informally, we say the training data exhibit \textbf{representation bias} when some subgroups are severely underrepresented~\cite{shahbazi2023representation}, such as ELL students and in particular high-scoring ELL students, who constitute only a tiny fraction of the data. As we show below, the losses from these subgroups contribute only weakly to the ERM objective and its gradient.

We view each example $(x,y,g)$ as drawn from a joint data distribution $P$.
For a given scoring function $f$, the population risk is
$R(f) = \mathbb{E}_{(x,y,g)\sim P}\bigl[\ell(f(x), y)\bigr]$.
Using the group label $g$, the population risk decomposes into a mixture of group-wise
risks:
\begin{equation}
  R(f)
    = \sum_{g \in \{\text{EL},\text{NE}\}}
         P(g)\,
         \mathbb{E}\bigl[\ell(f(x), y) \mid g\bigr]
    = \sum_{g \in \{\text{EL},\text{NE}\}} \pi_g\,R_g(f),
\end{equation}
where $\pi_g = P(g)$ is the population ratio of group $g$ and
$R_g(f) = \mathbb{E}\bigl[\ell(f(x), y) \mid g\bigr]$ is the group-wise risk.

Given the training set $D$ of size $n$ introduced in Section~\ref{sec:erm-problem}, the empirical risk in Equation~\eqref{eq:erm} admits a similar decomposition. Let $n_g = \bigl|\{ i \mid g_i = g \}\bigr|$ denote the number of examples from group $g$, and define the empirical group-wise risk
\begin{equation}
  \hat{R}_g(f_\theta)
    = \frac{1}{n_g}
      \sum_{i=1}^{n}
        \mathbbm{1}\{g_i = g\} \ell\bigl(f_\theta(x_i), y_i\bigr).
 \end{equation}
Then
\begin{equation}
  \hat{R}(f_\theta)
    = \frac{1}{n} \sum_{i=1}^{n}
         \ell\bigl(f_\theta(x_i), y_i\bigr) \\
    = \sum_{g \in \{\text{EL},\text{NE}\}}
         \frac{n_g}{n}\,\hat{R}_g(f_\theta).
\end{equation}

The gradient of the ERM objective decomposes accordingly:
\begin{equation}
  \label{eq:gradient}
  \nabla_\theta \hat{R}(f_\theta)
    = \sum_{g \in \{\text{EL},\text{NE}\}}
        \frac{n_g}{n}\,
        \nabla_\theta \hat{R}_g(f_\theta).
\end{equation} 
Under i.i.d. sampling, $n_g/n$ is a consistent estimator of $\pi_g$. Equation~\eqref{eq:gradient} shows that each group’s contribution to the ERM gradient is proportional to its sample fraction. When a group is underrepresented, its gradient term is naturally down-weighted, and optimization is driven primarily by the majority group. Similar observations have been made in prior work on fairness under ERM and distributional robustness~\cite{hashimoto2018fairness,leino2018feature}.
In our setting, this phenomenon is particularly severe for the high-scoring ELL subgroup $\mathcal{S}_{\text{EL,high}}$, which in our dataset constitutes only about 3.76\% of ELL responses and less than 1\% of all responses (see Section~\ref{sec:dataset}). As a result, its learning signal contributes negligibly to ERM optimization, and the high-score decision boundary is shaped largely by non-ELL examples.

\subsection{From Representation Bias to Bias Amplification}
\label{sec:rep-to-ba}

Having established that ERM down-weights learning signals from underrepresented subgroups, we now characterize how this imbalance can lead to bias amplification. We define \textbf{bias amplification} as the phenomenon in which the disparities between groups in model predictions exceed those present in the training data~\cite{wang2021directional,zhao2017men}.

To quantify bias amplification in automated scoring, we follow the idea of the directional bias amplification metric $BiasAmp_{\mathrm{\rightarrow}}$~\cite{wang2021directional} and focus on the direction $A\!\rightarrow\!T$, i.e., how the protected attribute $A$ affects the prediction of the task outcome $T$ beyond what is present in the labeled data. In our setting, the English group label $g \in \{\text{EL}, \text{NE}\}$ serves as the protected attribute $A$, and the task is to predict whether a student will get a high score, formalized as the binary indicator $T=\mathbbm{1}[y \ge \tau]$. Given a scoring model $f$, we obtain predicted scores $\hat{y} = f(x)$ and define the predicted high-score indicator $\hat{T} = \mathbbm{1}[\hat{y} \ge \tau]$. For each group $g \in \{\text{EL}, \text{NE}\}$, we define the high-score rate in the labeled data and in the model predictions as 
$p_{\text{true}}(g) = \Pr(T = 1 \mid g)$,
$p_{\text{pred}}(g) = \Pr(\hat{T} = 1 \mid g)$.

The \textbf{group disparity} in high-score rates in the labeled data and model is then
$\Delta_{\text{data}} = p_{\text{true}}(\mathrm{NE}) - p_{\text{true}}(\mathrm{EL})$, and 
$\Delta_{\text{model}} = p_{\text{pred}}(\mathrm{NE}) - p_{\text{pred}}(\mathrm{EL})$, respectively.
We define our bias amplification metric as the change in this disparity induced by the model training process: 
\begin{align}
\mathrm{BiasAmp}(f)
&= \Delta_{\text{model}} - \Delta_{\text{data}}.
\end{align}
Here $\mathrm{BiasAmp}(f) > 0$ means that the model amplifies the bias in the same direction as in the labeled data. 
$\mathrm{BiasAmp}(f) \approx 0$ indicates that the model largely preserves the disparity present in the labeled distribution, whereas $\mathrm{BiasAmp}(f) < 0$ corresponds to attenuation or even
reversal of this disparity.

To theoretically substantiate how representation bias leads to bias amplification, we decompose $\mathrm{BiasAmp}(f)$ into its constituent classification error terms. For each group $g$, define the joint false-positive and false-negative rates
$\mathrm{FP}_g = \Pr(T = 0, \hat{T} = 1 \mid g)$,
$\mathrm{FN}_g = \Pr(T = 1, \hat{T} = 0 \mid g)$.
Within a fixed group $g$, the true and predicted high-score rates can be expressed in terms of these quantities:
\begin{align}
p_{\text{true}}(g)
  &= \Pr(T = 1 \mid g)
   = \Pr(T = 1, \hat{T} = 1 \mid g) + \mathrm{FN}_g, \\
p_{\text{pred}}(g)
  &= \Pr(\hat{T} = 1 \mid g)
   = \Pr(T = 1, \hat{T} = 1 \mid g) + \mathrm{FP}_g.
\end{align}
Hence
$p_{\text{pred}}(g) - p_{\text{true}}(g) = \mathrm{FP}_g - \mathrm{FN}_g$.
Intuitively, $\mathrm{FP}_g$ is the fraction of low-scoring responses in group $g$ that the model
``upgrades'' to high scores (overpredictions), whereas $\mathrm{FN}_g$ is the fraction of truly
high-scoring responses that it ``downgrades'' (underpredictions). Thus
$p_{\text{pred}}(g) - p_{\text{true}}(g)$ is exactly the net upward shift in the high-score rate
induced by the model for group $g$.

Substituting $p_{\text{pred}}(g) - p_{\text{true}}(g) = \mathrm{FP}_g - \mathrm{FN}_g$ for
$g \in \{\mathrm{NE}, \mathrm{EL}\}$ into the definition of $\mathrm{BiasAmp}(f)$ yields
\begin{align}
\mathrm{BiasAmp}(f)
&= \bigl[\mathrm{FP}_{\mathrm{NE}} - \mathrm{FN}_{\mathrm{NE}}\bigr]
 - \bigl[\mathrm{FP}_{\mathrm{EL}} - \mathrm{FN}_{\mathrm{EL}}\bigr] \\
&= \bigl[\mathrm{FP}_{\mathrm{NE}} - \mathrm{FP}_{\mathrm{EL}}\bigr]
 + \bigl[\mathrm{FN}_{\mathrm{EL}} - \mathrm{FN}_{\mathrm{NE}}\bigr].
\end{align}
This decomposition shows that \textbf{bias amplification arises} when the model assigns more false-positive high scores to non-ELLs (\(\mathrm{FP}_{\mathrm{NE}} > \mathrm{FP}_{\mathrm{EL}}\)), while producing more false-negative errors for ELLs (\(\mathrm{FN}_{\mathrm{EL}} > \mathrm{FN}_{\mathrm{NE}}\)).
This pattern is expected under representation bias and ERM-dominated training.

Under the representation bias characterized in Section~\ref{sec:erm-imbalance}, \textbf{the high-score decision boundary is shaped predominantly by non-ELL responses}~\cite{hashimoto2018fairness}. 
During training, the model may conflate construct-relevant content with linguistic patterns typical of high-scoring non-ELL responses, effectively treating these group-specific patterns as a proxy for high proficiency. In contrast, high-scoring ELL responses often exhibit valid but sparsely represented linguistic features~\cite{crossley2012predicting,weigle2013english}, so the model tends to under-predict high-scoring ELL responses at test time~\cite{hashimoto2018fairness}. 
This mechanism naturally induces \(\mathrm{FP}_{\mathrm{NE}} > \mathrm{FP}_{\mathrm{EL}}\) and \(\mathrm{FN}_{\mathrm{EL}} > \mathrm{FN}_{\mathrm{NE}}\), and, combined with the decomposition above, implies \(\mathrm{BiasAmp}(f) > 0\).

\section{Mitigating Bias Amplification with BRIDGE} 

Our analysis in Section~\ref{sec:biasamp} demonstrates that bias amplification arises from skewed decision boundaries under ERM. Left unaddressed, such biased predictions risk perpetuating a feedback loop (Fig.~\ref{architecture}(a)) in which underprediction discourages minority students, leading to behavioral and content production biases that further reinforce representation bias.
To break this cycle, we propose BRIDGE, a framework designed to intervene at the data level by augmenting the sparse $\mathcal{S}_{\text{EL,high}}$ subgroup.
BRIDGE (Fig.~\ref{architecture}(c)) operates in two stages: (i) an LLM-based generator performs inter-group stylistic reformulation to ``paste'' high-score content into ELL linguistic patterns; (ii) a discriminator filters synthetic candidates to ensure they are indistinguishable from authentic student responses.

\subsection{Stage 1: Inter-group Stylistic Reformulation}
The primary objective of the first stage is to synthesize high-quality responses that occupy the sparse $\mathcal{S}_{\text{EL,high}}$ region.
While high-scoring ELL samples are rare, the complementary abundance of high-scoring non-ELL content and representative ELL linguistic patterns offers a viable path for synthesis.
However, effectively leveraging these resources requires a principled decoupling justified by construct validity, which holds that domain-specific assessments should primarily target construct-relevant proficiency while minimizing construct-irrelevant variance arising from extraneous linguistic barriers~\cite{abedi2006psychometric,AERA2014Standards,messick1987validity,zhai2020evaluation}.
Motivated by this principle, we conceptualize a student response $x$ as a modular composition of construct-relevant content $C(\cdot)$ and linguistic style $S(\cdot)$.

Empirical analyses of five datasets support this conceptualization. Even at the same score level, ELL responses $x_{\text{EL}}$ and non-ELL responses $x_{\text{NE}}$ remain linearly discriminable in the BERT latent space~\cite{devlin2019bert} (ROC AUC: 0.794--0.830).
Crucially, identical scores are intended to reflect comparable content performance between ELL and non-ELL, and any residual content variations would typically average out across large samples. Thus, the persistent discriminability we observe suggests that the distinction resides not in the construct-relevant content, but in group-specific stylistic distributions.
Building on this, BRIDGE formalizes the synthesis of high-quality ELL samples as a conditional generation task:
\begin{equation}
x_{\text{{syn}}} = G\big(C(x_{\text{NE}}), S(x_{\text{EL}})\big),
\end{equation}
where $x_{\text{NE}} \in \mathcal{S}_{\text{NE,hi}}$ is the content donor and $x_{\text{EL}}$ is drawn from the ELL population. Specifically, the generator $G$ is instructed to ``paste'' the construct-relevant content of $x_{\text{NE}}$ into the authentic ELL linguistic patterns of $x_{\text{EL}}$, bridging the representation gap without introducing construct-irrelevant bias.

\subsection{Stage 2: Discriminative Filtering for Authenticity}
While Stage 1 reformulates high-score content into ELL patterns, outputs may still contain artifacts that deviate from the authentic distribution. 
Thus, we introduce a Discriminative Filtering stage to ensure authenticity of the augmented data.
Specifically, we train a discriminator $D$ to distinguish between authentic responses and the synthetic candidates generated in Stage 1. For each candidate $x_{\text{syn}}$, the discriminator outputs an authenticity score $D(x_{\text{syn}}) \in [0, 1]$, representing the probability that the response originates from a real student rather than an LLM. 
We retain samples that exceed a predefined confidence threshold $\gamma$:
\begin{equation}
\mathcal{S}_{\text{syn}}^* = \{\, x_{\text{syn}} \mid D(x_{\text{syn}}) > \gamma \,\}.
\end{equation}
By filtering out overly polished or hallucinated responses, this stage helps the augmented set $\mathcal{S}_{\text{syn}}^*$ better match the linguistic profile of the ELL population while providing the necessary high-score evidence.

\section{Experiments}
In this section, we evaluate the performance of the BRIDGE framework across five diverse science assessment items. Our experiments are guided by the following three research questions: 
\textbf{RQ1:} Does the ERM-based scoring model amplify prediction disparity for the sparse $\mathcal{S}_{\text{EL,high}}$ subgroup?
\textbf{RQ2:} To what extent can BRIDGE reduce this bias amplification compared to existing augmentation methods?
\textbf{RQ3:} Can BRIDGE maintain competitive overall scoring accuracy while improving fairness for the $\mathcal{S}_{\text{EL,high}}$ subgroup?

\subsection{Datasets}
\label{sec:dataset}

Following prior work~\cite{guo2025artificial}, we use Grade~8 student constructed-response data from the California Science Test (CAST), a state-administered assessment of science knowledge. Students write short answers that are scored by human raters on a 0/1/2 scale. We collapse the original seven English proficiency categories into a binary group label: the ELL group and the Non-ELL group. Across items, ELL students account for about 14.2\% of responses (range $\approx$ 11.7\%–17.0\%).

\noindent\textbf{The Sparsity Challenge.} 
We focus on the intersectional subgroup $\mathcal{S}_{\text{EL,high}}$ (ELL students achieving score 2). This subgroup is rare, accounting for a median of 3.76\% (mean = 8.28\%) within the ELL population. In typical small to medium items containing about 200 ELL responses, about 60\% of the items contain 10 or fewer $\mathcal{S}_{\text{EL,high}}$ samples. Consequently, a standard train/validation/test split (e.g., 60/20/20) would yield a test set with only a few $\mathcal{S}_{\text{EL,high}}$ samples (typically $\leq 3$ samples), making subgroup metrics statistically unreliable or even undefined.

\noindent\textbf{Simulated Low-Resource Scenario.} 
To investigate extreme sparsity during training while maintaining reliable evaluation, we select five large-scale items, each containing roughly 200,000 responses (30,000 ELL responses). For these items, the absolute count of $\mathcal{S}_{\text{EL,high}}$ samples is sufficient for testing (ranging from $713$ to $3,658$), while their within-ELL proportions mirror the sparse distributions of smaller items. For each item, we perform a two-stage stratified random split. Data are first divided into training ($70\%$), validation ($15\%$), and test ($15\%$) sets. We then subsample $1\%$ of the training data to simulate an extreme low-resource scenario. The remaining $99\%$ of the training pool is reserved solely as a donor source for real-data (oracle) augmentation experiments. The validation and test sets remain fixed across all conditions to ensure reliable performance estimation.

\subsection{Experimental Setup}

\noindent\textbf{Scoring Models.} 
We fine-tune \texttt{bert-base-uncased} as a three-way sequence classifier over score level 0/1/2, with a maximum input length of $256$ tokens. The model is trained with cross-entropy loss and optimized using AdamW (learning rate $2\times10^{-5}$, weight decay $0.01$, batch size $16$) and a linear learning-rate schedule with warm-up for up to 20 epochs. We employ early stopping with a patience of 3 epochs (minimum improvement: 0.001), retaining the checkpoint with the highest validation accuracy for test-set evaluation. All experiments are repeated over 10 random seeds to ensure statistical stability.

\noindent\textbf{BRIDGE Implementation.} 
To synthesize additional $\mathcal{S}_{\text{EL,high}}$ samples, we utilize GPT-4o~\cite{hurst2024gpt} ($T=0.8$) to perform the inter-group stylistic reformulation on the training split.  Following the framework described in Section 3, we extract construct-relevant content from non-ELL high-scoring responses and reformulate it using ELL-specific linguistic patterns. Candidates are then passed through a lightweight discriminator trained on a mixture of authentic and synthetic responses for up to 10 epochs. We retain only synthetic samples exceeding an authenticity threshold of $\gamma = 0.6$, which are assigned the $\mathcal{S}_{\text{EL,high}}$ label and used as augmented training data in the subsequent experiments.

\noindent\textbf{Baselines.} 
We compare BRIDGE against three augmentation strategies while keeping the scoring architecture and optimization constant.
(i) \emph{Real-data augmentation (Oracle):} Adds additional authentic $\mathcal{S}_{\text{EL,high}}$ samples, drawn from the held-out 99\% pool, to the 1\% training set up to a fixed budget. This serves as an Oracle upper bound, representing the model's potential when provided with additional real human data.
(ii) \emph{Oversampling:} A standard baseline for addressing class imbalance. Randomly replicates existing $\mathcal{S}_{\text{EL,high}}$ samples within the $1\%$ training set with replacement to meet the budget without introducing new linguistic content.
(iii) \emph{Paraphrasing}: An intra-group baseline using GPT-4o to rewrite available $\mathcal{S}_{\text{EL,high}}$ samples, relying solely on existing minority patterns without leveraging external content.
All baselines differ exclusively in the construction of the additional $\mathcal{S}_{\text{EL,high}}$ training instances, ensuring observed performance gains are attributable to the augmentation strategy.

\noindent\textbf{Evaluation Metrics.} 
We evaluate models across two dimensions: scoring performance and group fairness. 
Scoring performance is measured by \emph{Accuracy}, \emph{Quadratic Weighted Kappa (QWK)}, \emph{Mean Absolute Error (MAE)}, and \emph{Macro-F1}, providing a holistic view of agreement and robustness. 
To quantify fairness, we report: 
(i) \emph{MSG Gap:} Defined as the difference between the model’s Non-ELL vs.\ ELL Mean Score Gap (MSG) and the corresponding MSG under human scores. Positive values indicate that the model widens the human MSG.
(ii) \emph{BiasAmp:} As defined in Section 2.3, this metric measures bias amplification for the $\mathcal{S}_{\text{EL,high}}$ subgroup by capturing the change in the Non-ELL vs.\ ELL disparity at the score 2 level between human and model scores.
All metrics are reported as mean $\pm$ standard deviation over 10 random seeds.

\subsection{Results and Analysis}

\begin{table}[t]
\centering
\caption{Bias mitigation performance of different data augmentation methods on five CAST datasets, each with 20 additional $\mathcal{S}_{\text{EL,high}}$ samples per method.
Mean $\pm$ standard deviation over 10 random seeds. Best and second-best MSG Gap and BiasAmp are highlighted in \textbf{bold} and \underline{underlined}, respectively.}
\newcommand{\std}[1]{{\scriptsize$\pm$#1}}
\newcommand{\best}[1]{\textbf{#1}}
\newcommand{\second}[1]{\underline{#1}}
\setlength{\tabcolsep}{2pt}
\renewcommand{\arraystretch}{1.05}

\resizebox{\columnwidth}{!}{%
\begin{tabular}{l l c c c c c c}
\toprule
& & \multicolumn{2}{c}{Fairness} & \multicolumn{4}{c}{Performance} \\
Dataset & Aug. Method & MSG Gap$\downarrow$ & BiasAmp$\downarrow$ & Acc.(\%)$\uparrow$ & MAE$\downarrow$ & QWK(\%)$\uparrow$ & F1(\%)$\uparrow$ \\
\cmidrule(r){3-4} \cmidrule(l){5-8}
\midrule
Dataset 1 & Baseline (no aug.)       
          & 0.0279\std{0.0134} & 0.0097\std{0.0081} & 90.17\std{0.48} & 0.0988\std{0.0053} & 89.77\std{0.64} & 88.68\std{0.50} \\
          & + Real-data (oracle)    
          & \second{0.0100}\std{0.0129} & \second{0.0025}\std{0.0093} & 90.19\std{0.48} & 0.0993\std{0.0046} & 89.68\std{0.34} & 88.58\std{0.59} \\
          & + Oversampling 
          & 0.0279\std{0.0092} & 0.0107\std{0.0054} & 90.41\std{0.42} & 0.0962\std{0.0043} & 90.05\std{0.40} & 89.01\std{0.40} \\
          & + Paraphrasing   
          & 0.0268\std{0.0137} & 0.0082\std{0.0094} & 90.31\std{0.32} & 0.0974\std{0.0035} & 89.86\std{0.39} & 88.77\std{0.33} \\
          & + Ours         
          & \best{0.0079}\std{0.0139}  & \best{-0.0014}\std{0.0085} & 90.22\std{0.27} & 0.0994\std{0.0031} & 89.53\std{0.38} & 88.63\std{0.27} \\
\midrule
Dataset 2 & Baseline (no aug.)       
          & 0.0125\std{0.0045} & 0.0132\std{0.0039} & 94.32\std{0.32} & 0.0568\std{0.0032} & 96.15\std{0.21} & 92.81\std{0.41} \\
          & + Real-data (oracle)    
          & \second{0.0120}\std{0.0089} & \best{0.0088}\std{0.0072}  & 94.38\std{0.25} & 0.0564\std{0.0024} & 96.17\std{0.16} & 92.86\std{0.33} \\
          & + Oversampling 
          & 0.0145\std{0.0056} & 0.0121\std{0.0038} & 94.48\std{0.18} & 0.0553\std{0.0018} & 96.26\std{0.12} & 92.99\std{0.26} \\
          & + Paraphrasing   
          & 0.0155\std{0.0056} & 0.0132\std{0.0024} & 94.32\std{0.32} & 0.0569\std{0.0032} & 96.15\std{0.21} & 92.78\std{0.41} \\
          & + Ours         
          & \best{0.0094}\std{0.0086}  & \second{0.0097}\std{0.0056} & 94.39\std{0.41} & 0.0562\std{0.0042} & 96.18\std{0.27} & 92.89\std{0.62} \\
\midrule
Dataset 3 & Baseline (no aug.)       
          & 0.0032\std{0.0092} & 0.0073\std{0.0066} & 91.28\std{0.32} & 0.0874\std{0.0033} & 93.43\std{0.20} & 87.76\std{0.51} \\
          & + Real-data (oracle)    
          & \second{-0.0034}\std{0.0183} & \second{0.0034}\std{0.0130} & 91.27\std{0.39} & 0.0876\std{0.0039} & 93.39\std{0.34} & 87.83\std{0.53} \\
          & + Oversampling 
          & -0.0006\std{0.0156} & 0.0036\std{0.0111} & 91.41\std{0.29} & 0.0861\std{0.0030} & 93.47\std{0.27} & 88.04\std{0.40} \\
          & + Paraphrasing   
          & 0.0033\std{0.0130} & 0.0075\std{0.0109} & 91.38\std{0.33} & 0.0864\std{0.0033} & 93.48\std{0.31} & 87.93\std{0.47} \\
          & + Ours         
          & \best{-0.0037}\std{0.0117} & \best{0.0031}\std{0.0080} & 91.24\std{0.28} & 0.0880\std{0.0026} & 93.40\std{0.20} & 87.70\std{0.46} \\
\midrule
Dataset 4 & Baseline (no aug.)       
          & 0.0140\std{0.0225} & \best{0.0156}\std{0.0126} & 88.71\std{0.60} & 0.1131\std{0.0060} & 92.79\std{0.41} & 86.32\std{0.71} \\
          & + Real-data (oracle)    
          & \best{0.0111}\std{0.0143} & \second{0.0170}\std{0.0092} & 88.78\std{0.48} & 0.1126\std{0.0047} & 92.84\std{0.26} & 86.23\std{0.83} \\
          & + Oversampling 
          & 0.0172\std{0.0140} & 0.0200\std{0.0069} & 89.00\std{0.43} & 0.1104\std{0.0042} & 92.97\std{0.23} & 86.58\std{0.74} \\
          & + Paraphrasing   
          & 0.0176\std{0.0154} & 0.0223\std{0.0069} & 88.70\std{0.39} & 0.1133\std{0.0037} & 92.79\std{0.22} & 86.19\std{0.66} \\
          & + Ours         
          & \second{0.0128}\std{0.0157} & \second{0.0170}\std{0.0050} & 88.92\std{0.44} & 0.1111\std{0.0043} & 92.93\std{0.27} & 86.45\std{0.55} \\
\midrule
Dataset 5 & Baseline (no aug.)       
          & 0.0134\std{0.0154} & 0.0135\std{0.0102} & 91.46\std{0.29} & 0.0865\std{0.0030} & 94.25\std{0.21} & 88.35\std{0.54} \\
          & + Real-data (oracle)    
          & 0.0122\std{0.0088} & 0.0131\std{0.0043} & 91.41\std{0.29} & 0.0868\std{0.0029} & 94.22\std{0.22} & 88.43\std{0.45} \\
          & + Oversampling 
          & \second{0.0113}\std{0.0081} & \second{0.0119}\std{0.0054} & 91.51\std{0.26} & 0.0858\std{0.0027} & 94.32\std{0.16} & 88.37\std{0.51} \\
          & + Paraphrasing   
          & 0.0194\std{0.0064} & 0.0142\std{0.0027} & 91.42\std{0.28} & 0.0870\std{0.0029} & 94.23\std{0.19} & 88.22\std{0.62} \\
          & + Ours         
          & \best{0.0009}\std{0.0173} & \best{0.0050}\std{0.0087} & 91.36\std{0.27} & 0.0878\std{0.0031} & 94.12\std{0.25} & 88.25\std{0.53} \\
\bottomrule
\end{tabular}}

\vspace{1 mm}
    \begin{minipage}{\linewidth}
        \footnotesize{\textit{Note.} F1 denotes the macro-averaged F1 score over score levels.}
    \end{minipage}
    
\label{maintable}
\end{table}

\subsubsection{Quantification of Bias Amplification (RQ1)}
To evaluate RQ1, we examine whether the vanilla scoring model (baseline) amplifies the Non-ELL vs.\ ELL disparity for the high-scoring subgroup $\mathcal{S}_{\text{EL,high}}$. For each dataset, we compute the high-score disparity in the training data, $\Delta_{\text{data}}$, and in the model predictions, $\Delta_{\text{model}}$, and obtain the bias amplification measure $\mathrm{BiasAmp} = \Delta_{\text{model}} - \Delta_{\text{data}}$ defined in Section~\ref{sec:rep-to-ba}. As shown in Table~\ref{maintable}, across the five datasets, the baseline model consistently exhibits positive bias amplification: $\mathrm{BiasAmp}$ ranges from $0.0073$ to $0.0156$ (mean = $0.0119$). In other words, the model further widens the between-group disparity in the probability of receiving the top score by about 3.4--9.0\% relative to the disparity already present in the training data.

To ensure these findings are statistically meaningful rather than artifacts of random initialization, we perform significance tests over all runs for each dataset. We test the null hypothesis $H_0: \mathrm{BiasAmp} = 0$ against the one-sided alternative $H_1: \mathrm{BiasAmp} > 0$. Following normality checks, we apply either one-sample $t$-tests or Wilcoxon signed-rank tests as appropriate. Across all five datasets, the baseline model consistently exhibits significant bias amplification (all $p < 0.01$). These results provide robust evidence that scoring models systematically amplify existing group disparities for the $\mathcal{S}_{\text{EL,high}}$ subgroup under data sparsity.

\subsubsection{Comparative Analysis of Mitigation Strategies (RQ2)}
To address RQ2, we compare the mitigation efficacy of BRIDGE against all baseline strategies. As summarized in Table~\ref{maintable}, BRIDGE reduces both group-level disparity (MSG Gap) and high-score amplification (BiasAmp) across the majority of datasets. For instance, in Dataset~5, BRIDGE reduces the MSG Gap from 0.0134 to 0.0009, a 93.28\% reduction in group-level disparity. Similarly, in Dataset~1, our method reduces the BiasAmp from 0.0097 to -0.0014, indicating that the model no longer under-predicts high-scoring ELLs. For datasets with larger initial bias amplification (e.g., Datasets~1 and 5), these reductions are statistically significant ($p < 0.01$), whereas for items with smaller baseline bias (e.g., Dataset~3), the numerical improvements are directionally consistent but statistically \mbox{inconclusive}.

Beyond these general improvements, BRIDGE demonstrates competitive performance against the Real-data aug. (Oracle), achieving numerically superior fairness metrics in Datasets~1, 3, and 5. For instance, in Dataset~5, BRIDGE achieves a BiasAmp of $0.0050$, representing a $61.83\%$ further reduction in high-score bias compared to the oracle ($0.0131$). We attribute this superiority to the fact that BRIDGE creates controlled examples by injecting prototypical Non-ELL high-score logic into ELL linguistic patterns. This targeted intervention helps the model learn to decouple surface-level language features from content quality, providing a more concentrated debiasing signal than the naturally occurring variations in authentic ELL samples.

Compared to intra-group baselines, BRIDGE exhibits crucial robustness in challenging scenarios. Oversampling and Paraphrasing show marginal or even negative effects across most datasets, confirming that intra-group augmentation lacks sufficient signal diversity when the target subgroup is sparse. Notably, in the most challenging scenario, Dataset~4, Oversampling and Paraphrasing yield ``adverse effects'', increasing BiasAmp by $28.21\%$ and $42.95\%$ respectively. In contrast, BRIDGE maintains comparable performance to the oracle baseline.

\subsubsection{Trade-off between Fairness and Scoring Performance (RQ3)}
Regarding RQ3, Table~\ref{maintable} demonstrates that the substantial fairness gains of BRIDGE do not compromise overall scoring performance. Across all five datasets, BRIDGE introduces negligible changes to all performance metrics: Accuracy varies by $-0.11\%$ to $+0.24\%$, QWK by $-0.27\%$ to $+0.15\%$, Macro F1 by $-0.11\%$ to $+0.15\%$, and MAE by $-1.77\%$ to $+1.50\%$. These variations fall within the standard deviation of the baseline model, indicating that BRIDGE effectively mitigates bias amplification without degrading the model's scoring capability. This suggests that the fairness-performance tradeoff, a common bottleneck in bias mitigation methods, is not a significant concern for BRIDGE.

\subsection{Ablation Study}
To investigate the contribution of the Discriminative Filtering (Stage~2), we compare the full BRIDGE framework against a ``generation-only'' variant (w/o Stage~2). Our results indicate that Stage~1 (Inter-group Stylistic Reformulation) serves as the primary driver for bias mitigation, while Stage~2 provides critical quality control and robustness gains. For instance, in Dataset~5, Stage~1 alone achieves a $90.78\%$ reduction in MSG Gap (from $0.0134$ to $0.0012$), with Stage~2 contributing a marginal further improvement to $0.0009$, yielding an additional absolute improvement of 0.0003 (about 2.5\% of the baseline). The necessity of Stage~2 becomes more apparent in scenarios where Stage~1's initial gains are constrained. For Dataset~1, Stage~1 reduces the MSG Gap by $44.90\%$ (from 0.0279 to 0.0154), but Stage~2 is required to reach the final $0.0079$ level, representing an additional $26.88\%$ reduction relative to the baseline. In Dataset~3, Stage~1 alone yields negligible improvements to BiasAmp ($0.0073 \rightarrow 0.0071$). Manual spot-checks suggest that the proportion of high-quality synthetic samples in Dataset~3 is lower than in other items. Consequently, the noise from low-fidelity outputs counteracts the augmentation signal. By filtering out such low-quality samples, Stage~2 enables a significant reduction in BiasAmp to $0.0031$. In saturated scenarios like Dataset~2, the marginal impact of Stage~2 (a delta of only $0.0001$ in MSG Gap and 0.0003 in BiasAmp) confirms its role as a reliability mechanism rather than a primary signal generator. Overall, these findings support our two-stage design: Stage~1 establishes the debiasing foundation, while Stage~2 ensures stylistic authenticity and stability.

\section{Conclusion and Discussion}
This work reveals and quantifies the phenomenon of bias amplification in ERM-based automated scoring models. To mitigate this, we propose BRIDGE, an inter-group data augmentation framework designed for extreme low-resource scenarios. By ``pasting'' construct-relevant content from high-scoring non-ELL responses into authentic ELL linguistic patterns and employing a discriminative filter to eliminate low-fidelity candidates, BRIDGE effectively fills the gaps in the training data distribution. Our experiments on real-world CAST datasets demonstrate that BRIDGE significantly reduces both group-level disparities and high-score bias amplification while maintaining overall scoring performance. Notably, BRIDGE achieves fairness performance comparable to augmentation using additional real data. This suggests that in practical settings where high-scoring ELL responses are inherently difficult to obtain, educational institutions can achieve bias mitigation without the prohibitive cost of collecting rare samples. These results highlight BRIDGE as a scalable, controllable, and cost-effective solution for ensuring algorithmic fairness in large-scale educational assessments.

Despite its efficacy, BRIDGE relies on the disentanglement of content and style, which may not fully hold in assessments where linguistic expression is part of the core construct. Additionally, this study is limited to short-answer science items, and its generalizability to long essays remains to be tested. Future research will extend BRIDGE to other data scarce educational contexts, explore finer-grained stylistic control, and investigate synergies with other fairness-enhancing strategies to build a more robust and equitable scoring ecosystem.

\begin{credits}
\subsubsection{\ackname}
This work is supported by the Institute of Education Sciences (IES) under Grant No. R305C240010 and Grant No. R305A240356. The views and conclusions expressed in this paper are those of the authors and do not necessarily reflect the views of the funding agencies.

\subsubsection{\discintname}
The authors have no competing interests to declare that are relevant to the content of this article.
\end{credits}

%
\bibliographystyle{splncs04}

\end{document}